\documentclass[10pt, a4paper]{article}
\usepackage{lrec-coling2024} 

\usepackage{graphicx} 
\usepackage{amssymb}
\usepackage{amsmath}
\usepackage{pifont}

\usepackage{ulem}
\usepackage{enumitem}

\title{Enhanced Coherence-Aware Network with Hierarchical Disentanglement for Aspect-Category Sentiment Analysis}

\name{Jin Cui\textsuperscript{1}, Fumiyo Fukumoto\textsuperscript{1}$^{\ast}$ \thanks{*Corresponding author.}, Xinfeng Wang\textsuperscript{1}, Yoshimi Suzuki\textsuperscript{1}, \\ \fontsize{12pt}{14pt}\selectfont
\textbf{Jiyi Li\textsuperscript{1}}, \textbf{Noriko Tomuro\textsuperscript{2}}, and \textbf{Wanzeng Kong\textsuperscript{3}}} 

\address{\textsuperscript{1} University of Yamanashi, \textsuperscript{2} DePaul University, \textsuperscript{3} Hangzhou Dianzi University \\
         \{g22dtsa5,fukumoto,g22dtsa7,ysuzuki,jyli\}@yamanashi.ac.jp\\
          tomuro@cs.depaul.edu, and kongwanzeng@hdu.edu.cn}

\abstract{
Aspect-category-based sentiment analysis (ACSA), which aims to identify aspect categories and predict their sentiments has been intensively studied due to its wide range of NLP applications. Most approaches mainly utilize intrasentential features. However, a review often includes multiple different aspect categories, and some of them do not explicitly appear in the review. Even in a sentence, there is more than one aspect category with its sentiments, and they are entangled intra-sentence, which makes the model fail to discriminately preserve all sentiment characteristics. In this paper, we propose an enhanced coherence-aware network with hierarchical disentanglement (ECAN) for ACSA tasks. Specifically, we explore coherence modeling to capture the contexts across the whole review and to help the implicit aspect and sentiment identification. To address the issue of multiple aspect categories and sentiment entanglement, we propose a hierarchical disentanglement module to extract distinct categories and sentiment features. Extensive experimental and visualization results show that our ECAN effectively decouples multiple categories and sentiments entangled in the coherence representations and achieves state-of-the-art (SOTA) performance. Our codes and data are available online:
\url{https://github.com/cuijin-23/ECAN}.
\\ \newline }

\begin{document}
\maketitleabstract
\section{Introduction}

With the rapid growth of Internet services, a large number of user-generated textual reviews have become available, which has drawn much attention to aspect-based sentiment analysis (ABSA) research \cite{zhang2022survey}. 
To date, many of its variants, including aspect-based sentiment classification (ABSC) and aspect term-based sentiment analysis (ATSA), have been studied \cite{zhang2020convolution,li2021dual,zhang2022ssegcn,ma2023amr}. One such attempt is aspect-category-based sentiment analysis (ACSA), which consists of two subtasks: aspect category detection (ACD), which identifies aspect categories, and aspect-category sentiment classification (ACSC), which predicts their sentiments~\cite{cai2020aspect,li2020multi,zhou2022semantic}. 

\begin{figure}[t]
\begin{center}
\includegraphics[width=\linewidth]{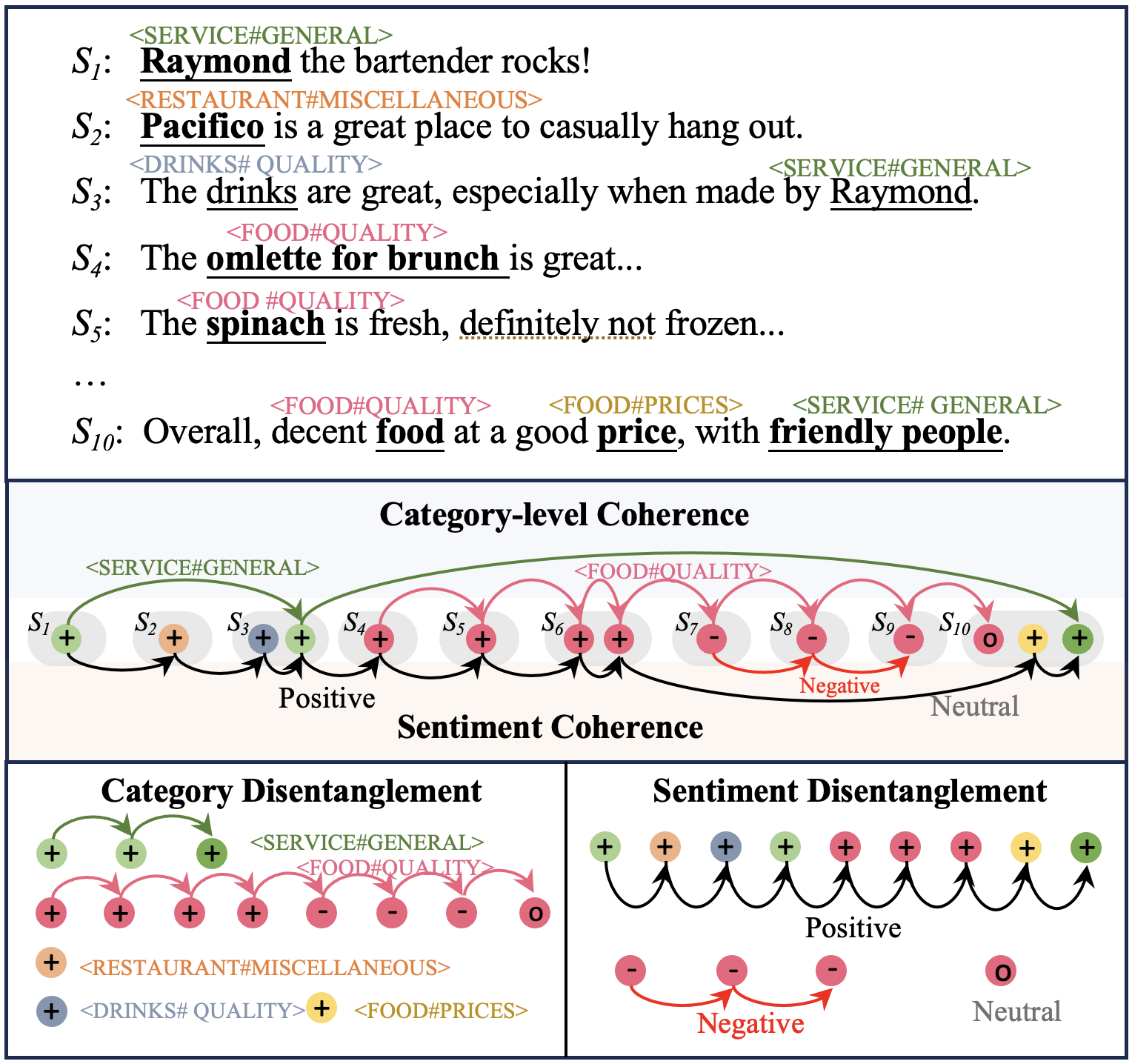} 
\caption{A review (ID: P\#9) from SemEval-2015: Words with angle brackets and the colored circle show aspect categories. ``$+$,'' ``$-$,'' and ``$\circ$'' denote positive, negative, and neutral sentiments, respectively.
Arcs indicate that they have an identical category or sentiment in a coherent review.}  
\label{motivation}
\end{center}
\end{figure}

The primary approaches to this task mainly leverage intrasentential features to learn their models~\cite{zhou2022semantic}. However, two major issues remain in the way of state-of-the-art (SOTA) performance. One is that a review often includes multiple different aspect categories, and some of them do not explicitly appear in the review. 
Taking an example from the REST 15 dataset shown in Figure~\ref{motivation}, there are several aspect categories, such as ``service'' and ``restaurant.'' ``Raymond'' in $S_3$ may be a bartender's name, while we cannot identify that it is related to the ``service'' category with only $S_3$, since ``service'' in $S_3$ shows an implicit aspect category. A user often gives reviews with a consistent opinion. For example, ``Raymond'' in $S_3$ is strongly related to both ``Raymond'' in $S_1$ and "friendly people'' in $S_{10}$ which indicates ``service'' served by people. However, the aforementioned approaches cannot capture how sentences are connected or how the entire review is organized to convey information to the reader.

Another issue is that there is more than one aspect category and its sentiment in a sentence, and they are entangled. A vanilla sentiment-level token fails to discriminately preserve all sentiment characteristics. For instance, $S_{10}$  in Figure \ref{motivation} contains a neutral opinion on the aspect of food, i.e., ``decent food'' and a positive opinion on the restaurant's price and service, i.e., ``good'' and ``friendly.'' Therefore, ``food'' is likely to be incorrectly identified as positive. As illustrated in the middle box in Figure~\ref{motivation}, i.e., category-level coherence, this indicates that both aspect categories and their sentiments are entangled intra-sentence as well as throughout sentences in a review.

Motivated by the issues mentioned above, we propose an enhanced coherence-aware network with hierarchical disentanglement (ECAN) for the ACSA task. Specifically, we leverage coherence modeling to capture contexts across the whole review. It also enables the model to learn explicit opinions from contexts to help the implicit aspect and sentiment identification. Furthermore, to address the issue that aspect categories and sentiments are entangled, we propose a hierarchical disentanglement module. The module applies a parallel attention mechanism on coherence-aware representation to both multiple categories and aspects to extract distinct categories and sentiment features. 
As illustrated in the bottom boxes of Figure~\ref{motivation}, we disentangle (i) the category representations, and (ii) the words that represent sentiments according to their sentiment polarities by utilizing the coherence features. The sentence-level representation is thus divided into several groups with relative category and sentiment features. 

In summary, the main contributions of our approach are as follows: 

(1) We propose an ECAN that leverages coherence modeling to learn the explicit opinions from contexts to help the implicit aspect and sentiment identification for the ACSA task;

(2) We propose a hierarchical disentanglement of category and sentiment for mining more fine-grained features; 

(3) Extensive experiments and visualization results on four benchmark datasets have demonstrated that the ECAN outperforms SOTA methods in both ACD and ACSC tasks.  

\begin{figure*}[th]
\begin{center}
\includegraphics[width=\linewidth]{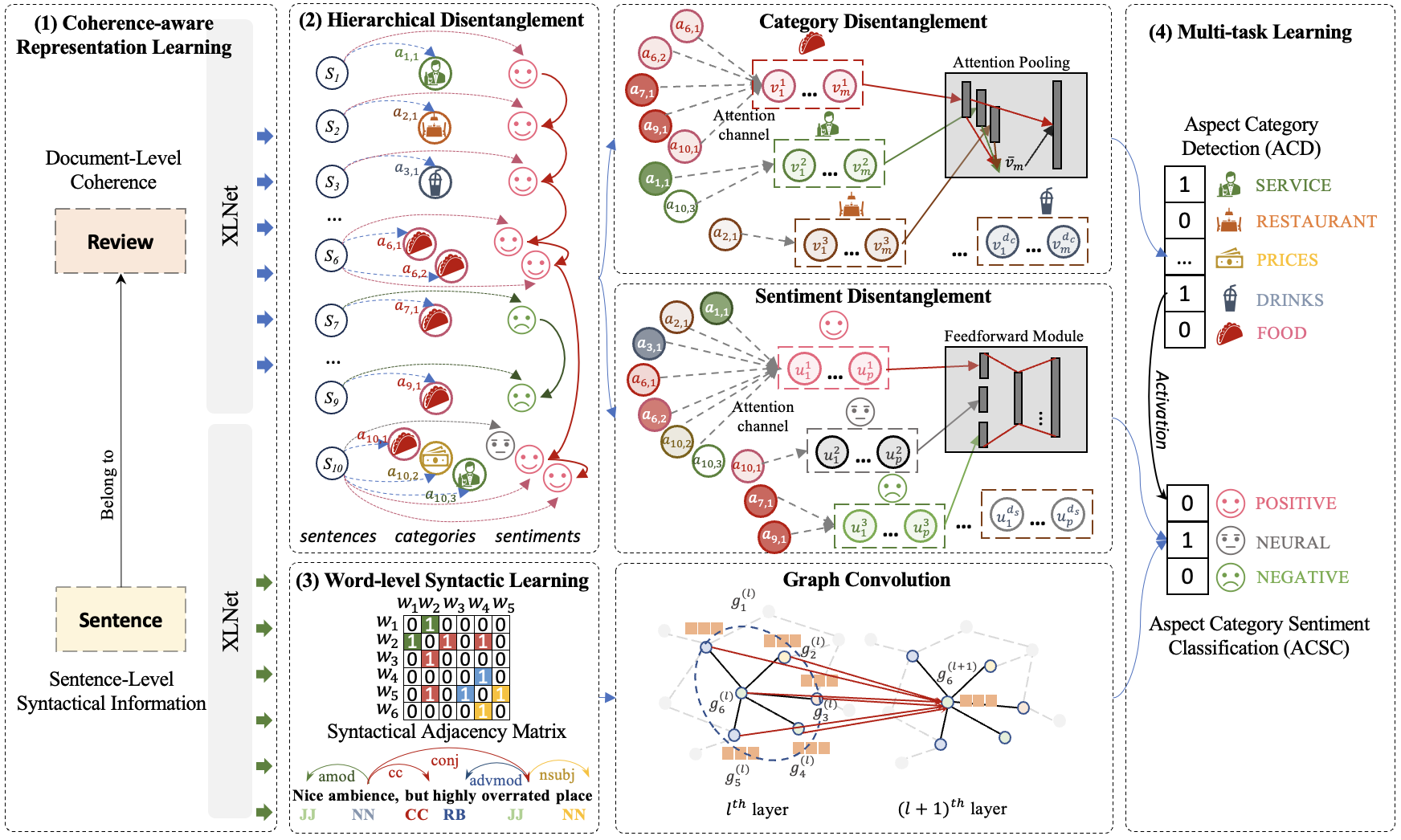} 
\caption{The main framework of our proposed method. It consists of four modules: (1) coherence-aware representation learning, (2) hierarchical disentanglement, (3) word-level syntactic learning, and (4) multi-task learning.} 
\label{main_framwork}
\end{center}
\end{figure*}

\section{Related Work}

\textbf{Aspect Category Sentiment Analysis}. Early studies on ACSA first detected the aspect category and then predicted the sentiment polarities for the detected categories~\cite{xue2018aspect,tay2018learning}. \citet{liang2019context} utilized a sparse coefficient vector to guide the representations of aspect category and sentiment. As such, the representations of aspect category and sentiment are learned independently, which hampers the performance of the model. Several studies focus on the issue and explore the joint prediction of two associated elements \cite{wang2019aspect,li2020joint,wu2021locate,lin2023capsule}, e.g., a multi-task learning approach~\cite{li2020multi} and a hierarchical graph convolutional network~\cite{cai2020aspect}. 

More recently, some attempts have been made to build sentiment triples or quadruples to simplify aspect-category tasks \cite{cai2021aspect,gao2022lego,gouetal2023mvp}. However, most of these approaches ignore the effectiveness of disentangling sentence representations by categories and sentiments in sentences.

\noindent
\textbf{Coherence Modeling}.
With the success of deep learning techniques, coherence analysis has been widely studied. Several approaches are designed to model sentence-level coherence, including a transferable neural model \cite{xu2019cross}, an entity-based neural local coherence model \cite{jeon2022entity},
and a local attention mechanism-based hierarchical coherence model \cite{liao2021hierarchical}.
Another approach is to learn coherence across a whole document, which includes a transformer-based architecture \cite{abhishek2021transformer} and a multi-task framework that learns both word- and document-level representations \cite{farag2019multi}. Our approach lies across both sentence- and document-level coherence to provide a comprehension framework for sentiment coherence.

\noindent
\textbf{Disentangled Representation Learning}.
Disentangled representation learning (DRL) has been widely employed to decouple underlying factors in the observable data \cite{bao2019generating,wang2022disentangled}, as it is intractable to model explanatory factors in some scenarios \cite{locatello2019challenging}.
Several DRL methods have been applied to NLP tasks, such as text-style
transfer \cite{cheng2020improving,john2019disentangled,nangi2021counterfactuals,hu2023token}, multi-modal sentiment analysis \cite{zhang2022learning}, domain adaptation \cite{liu2018detach,lee2021dranet,wang2023disentangled}, and text generation \cite{chen2019multi,thompson2020paraphrase}. 

Recent work on aspect-level sentiment analysis has primarily focused on the ATSA task. \citet{silva2021aspect} attempted to disentangle the syntactic and semantic features by utilizing the DeBERTa model \cite{he2020deberta}. 
\citet{mei2023disentangled} proposed a graph-based model that extracts a specific linguistic property to help capture finer feature representations. However, to date, little research has focused on aspect categories and sentiment disentanglement for the ACSA task.

\section{Task Definition}

Given a review $D$ with $I$ sentences, $D=\{s_i\}_{i=1}^I$, where the $i$-th sentence $s_i =\{ w_j\}_{j=1}^n$ consists of the number of $n$ words. 
Let also $C=\{c_i\}_{i=1}^m$ be a set of $m$ pre-defined aspect categories, and $p = \{ positive, neutral, negative \}$ be a set of sentiment polarity labels in $C$. The goal of the ACSA task is to detect all the aspect categories appearing in $s_i$ and classify the sentiment polarity for each detected category. 

\section{Approach}
We present an ECAN model consisting of four components, 
(1) coherence-aware representation learning with XLNet~\cite{yang2019xlnet},
(2) hierarchical disentanglement, (3) word-level syntactic learning for enhancing sentiment contexts, and (4) multi-task learning. The overall architecture is illustrated in Figure~\ref{main_framwork}. 

\subsection{Coherence-Aware Representation Learning with XLNet}

Unlike previous works \cite{yang2021improving,zhang2019aspect,zhang2022ssegcn} on ABSA which focus on local contexts within a sentence, ECAN mines both sentence- and document-level contexts from reviews to detect aspect categories and their sentiment polarities. To mine document-level contexts, we utilize the pre-training model XLNet to model the coherence representation. We choose the XLNet as the backbone, due to its capacity to learn longer text sequences, which has been widely used to model the coherence representation~\cite{cui2023aspect,jwalapuram2022rethinking,jeon2020centering}.

Specifically, for each document, we define an input sequence, $s_1$ [SEP] ... $s_I$ [SEP] [CLS], padded with two special tokens, [SEP] and [CLS], at the end of input $s_i$. Here, [SEP] and [CLS] are the same tokens as those of BERT \cite{devlin2018bert}. Then, we obtain the review embedding $e_d \in \mathbb{R}^{ I \times {d_m}}$ via XLNet. Likewise, for each sentence, we create an input, $s_i$ [SEP][CLS]; the word-level embedding is denoted as $ e_{w_{j}} \in \mathbb{R}^{{d_m}}$, where $d_m$ denotes the embedding size. 

To learn robust coherence representations, we employ the sentence ordering contrastive learning (CL) task ~\cite{jwalapuram2022rethinking} as an auxiliary task. It enforces that the coherence score of the positive sample (original document) should be higher than that of the negative sample (disorder document). We randomly shuffle the sentences within the original review to generate the number of $B$ negative samples and apply contrastive learning to align the coherent and incoherent representations. Let $f_{\theta} (\mathbf{e}_{d})$ be a linear projection to convert coherent document embedding $\mathbf{e}_d$ into coherence scores. The margin-based contrastive loss is given as follows:

 \begin{equation}
 \small{
 \mathcal{L}_{cl} = - \sum_{d^{+} \in \mathcal{D}} \log(\frac{\mathrm{exp}{(f_{\theta} ({\mathbf{e}_{d}}^+))}}{\mathrm{exp}({f_{\theta} ({\mathbf{e}_{d}}^+)}) + {\textstyle \sum_{ j=1}^{B}} \mathrm{exp}{(f_{\theta} ({\mathbf{e}_{d}}_j^-) - \tau )}}),
 }
\end{equation}

\noindent
where $d^{+}$ refers to a positive sample, $f_{\theta} ({\mathbf{e}_{d}}^+)$ indicates the score of the positive sample, $f_{\theta} ({\mathbf{e}_{d}}_1^-)$, ..., $f_{\theta} ({\mathbf{e}_{d}}_B^-)$ denote the scores of $B$ negative samples, and $\tau$ refers to the margin.

\subsection{Hierarchical Disentanglement}

Although coherence modeling with vanilla XLNet can capture contextual semantics, it fails to discriminate contextual features of different aspect categories and aspect sentiments within a sentence, as the coherence model captures global relationships among sentences within a whole document, which hinders the performance of more fine-grained aspect-based sentiment tasks, i.e., ACD and ACSC tasks. To address the issue, we simultaneously disentangle the categories and their sentiment representations within each sentence embedded by hierarchical document-level coherence.

\subsubsection{Category Disentanglement}

Inspired by the work of \citet{medina2018parallel}, we adopt a parallel attention mechanism to obtain disentangled representation blocks indicating fine-grained category components. Formally, for the $i$-th attention channel ($i \in \{1, 2, ..., d_c\}$), given the document-level representation $e_d$, we employ a slicing operation to obtain the representation $e_{s}$ of the sentence $s$ in $e_d$ as the $\mathbf{e}_i^{(0)}$ at the first layer, and then we stack transformer blocks to obtain the self-attention values in each channel:

\begin{equation}
\mathbf{e}_i^{(l+1)} = \mathrm{softmax}\left (\sum_{j \in \mathbf{s}_i} \frac{\mathbf{e}_i^{(l)} \mathbf{w}^{q_1}_{*,i}  (\mathbf{e}_j^{(l)} \mathbf{w}^{k_1}_{*,j})^\top}{\sqrt{d_k}} \right ) \mathbf{e}_i^{(l)} \mathbf{w}^{v_1}_{*,i},
\end{equation}

\noindent
where $\mathbf{W}^{q_1} \in \mathbb{R}^{d_m \times d_k}$, $\mathbf{W}^{k_1} \in \mathbb{R}^{d_m \times d_k}$ and $\mathbf{W}^{v_1} \in \mathbb{R}^{d_m \times d_k}$ are three weight matrices, and $d_k$ is the dimension size. Subsequently, $\mathbf{e}_i^{(l)}$ at the last layer is fed into the feedforward network as follows:

\begin{equation}
\mathbf{v}_i^{(l)} = \mathrm{ReLU}(\mathbf{e}^{(l)}_i\mathbf{w}_{*,i}^{FC_1}+\mathbf{b}_1)\mathbf{w}_{*,i}^{FC_2} + \mathbf{b}_2,
\end{equation}

\noindent
where $\mathbf{W}^{FC_1} \in \mathbb{R}^{d_m \times d_k}$, $\mathbf{W}^{FC_2} \in \mathbb{R}^{d_m \times d_k}$ are two linear transformation matrices, $\mathbf{b}_1$ and $\mathbf{b}_2$ are learnable parameters. After the number of $d_c$ parallel channels, we obtain the representations of $d_c$ disentangled blocks, i.e., $\mathbf{v}_1^{(l)},\mathbf{v}_2^{(l)},...,\mathbf{v}_{d_c}^{(l)}$. These disentangled representations are concatenated to further select representative features using the attention pooling layer by the following equation:

\begin{equation}
\begin{aligned}
M_j &= \mathrm{tanh}(\mathbf{W}_M^{\top}[\mathbf{v}_1^{(l)};\mathbf{v}_2^{(l)};...;\mathbf{v}_{d_c}^{(l)}]+\mathbf{b}_M), \\
\alpha_M &= \sum_{j \in d_j} \mathrm{softmax}(\mathbf{W}_j^{\top} M_j),
\end{aligned}
\end{equation}

\noindent
where $\mathbf{W}_M \in \mathbb{R}^{d_c \cdot d_m \times d_m}$, $\mathbf{W}_j \in \mathbb{R}^{d_m \times d_m}$ and $\mathbf{b}_M \in \mathbb{R}^{d_c \cdot d_m}$  are learnable parameters, and $\alpha_M$ is the attention weight vector.
The final category representation $r_c$ is denoted as:

\begin{equation}
r_c = [\mathbf{v}_1^{(l)};\mathbf{v}_2^{(l)};...;\mathbf{v}_{d_c}^{(l)}]\alpha_M.
\end{equation}

We note that the category representation $r_c$ is partially optimized by the ACD task, whose gradients are based on ground-truth category information.

\subsubsection{Sentiment Disentanglement}

Similar to category disentanglement, we obtain the disentangled sentiment representation blocks from the document-level coherence representation $e_d$ via a parallel attention mechanism, where the number of parallel channels is defined as $d_s$. The difference is that the disentangled sentiment representations are optimized via the ACSC task only, containing ground-truth sentiment information. We note that the gradient update of disentangled sentiment representations is completely trained by the ACSC task; although, whether it will be trained or not is activated by the aspect category information.

Specifically, in the $i$-th channel, the self-attention results are computed layer by layer as follows:

\begin{equation}
\small{
\mathbf{u}_i^{(l+1)} = \mathrm{softmax}\left (\sum_{j \in \mathbf{s}_i} \frac{\mathbf{u}_i^{(l)} \mathbf{w}^{q_2}_{*,i}  (\mathbf{u}_j^{(l)} \mathbf{w}^{k_2}_{*,j})^\top}{\sqrt{d_k}} \right ) \mathbf{u}_i^{(l)} \mathbf{w}^{v_2}_{*,i},
}
\end{equation}

\noindent
where $\mathbf{W}^{q_2}$, $\mathbf{W}^{k_2}$ and $\mathbf{W}^{v_2}$ are three weight matrices. Empirically, we found that a simple linear transformation works better than an attention-pooling operation.

\begin{equation}
\mathbf{U} = [\mathbf{u}^{(l)}_1,\mathbf{u}^{(l)}_2,...,\mathbf{u}^{(l)}_{d_s} ]\mathbf{W}_U.
\end{equation}

Thereafter, the output $\mathbf{U}$ is then fed into a position-wise feed-forward neural network, generating hidden disentangled representations $\mathbf{u}_s$ of the sentiment representation: 

\begin{equation}
\mathbf{u}_s = \mathrm{ReLU}(\mathbf{U}\mathbf{W}^{FC_1}+\mathbf{b}_1)\mathbf{W}^{FC_2} + \mathbf{b}_2,
\end{equation}

\noindent
where $\mathbf{W}^{FC_1}$, $\mathbf{W}^{FC_2}$ are two weight matrices.

\subsection{Word-Level Syntactic Learning for Enhancing Sentiment Semantics}

To capture the local contexts within a sentence, we use the Stanford parser\footnote{\url{https://stanfordnlp.github.io/CoreNLP/}} to obtain a word-level dependency tree of the target sentence and apply graph convolution operations to learn the local sentiment correlations between sentiment and aspect categories from the dependency tree. 
The syntactic adjacency matrix in the dependency tree of the sentence is defined by $\mathbf{A}\in \mathbb{R}^{n\times n}$, and the syntactic embedding nodes are denoted by $\mathbf{g}^{(0)}=[\mathbf{e}_{w_1},\mathbf{e}_{w_2},..., \mathbf{e}_{w_n}]$. Then, the normalized adjacency matrix can be obtained by $\hat{\mathbf{A}} = \mathbf{D}^{-\frac{1}{2}} \mathbf{A} \mathbf{D}^{-\frac{1}{2}}$, where $\mathbf{D} \in \mathbb{R}^{N \times N}$ is a diagonal matrix in which each entry, $d_{ij} = {\textstyle \sum_{j=1}^{n}}\mathbf{a}_{ij}$ denotes the number of nonzero entries in the $i$-th row vector of the adjacency matrix $\mathbf{A}$. The node representations are updated as follows:

\begin{equation}
\begin{aligned}
{\tilde{\mathbf{g}_i}}^{(l)} &= \sum_{j=1}^{n} \hat{\mathbf{a}}_{ij}\mathbf{W}^{(l)} \mathbf{g}^{(l-1)}_j, \\
\mathbf{g}^{(l)}_i &=  {\rm ReLU} (\tilde{\mathbf{g}}^{(l)}_i+\mathbf{b}^{(l)}),
\end{aligned}
\end{equation}

\noindent
where $\mathbf{g}^{(l-1)}_i \in \mathbb{R}^{d_m}$ denotes the $i$-th word representation obtained from the graph convolution network (GCN) layer, and $\mathbf{g}^{(l)}_i$ refers to the $i$-th word representation of the current GCN layer. The weights $ \mathbf{W}^{(l)}$ and bias $\mathbf{b}^{(l)}$ are learnable parameters. We feed the obtained local sentiment representation $\mathbf{g}^{(l)}_i$ into a linear layer, which can be denoted as $\hat{\mathbf{g}}\in \mathbb{R}^{d_m} $:

\begin{equation}
\hat{\mathbf{g}} = [\mathbf{g}^{(l)}_i,...,\mathbf{g}^{(l)}_n] \times \mathbf{W}_o^T + \mathbf{b}_o,
\end{equation}

\noindent
where $W_o$ and $b_o$ are the learnable weight and bias, respectively.

\subsection{Multi-Task Learning}
We use a multi-task framework to jointly learn to identify document-level coherence with sentence ordering contrastive learning, detect the underlying aspect categories, and identify their sentiments. 

\noindent
\textbf{Aspect-category detection}.
The final category representation is obtained by $r_c$; thus the probability of the $i$-th category $p_j^c \in \mathbb{R}^m $ is given by:

\begin{equation}
p^c_j = p(y_j^c|r_j) = \mathrm{sigmoid}(\mathbf{W}^c_j r_c+\mathbf{b}^c_j).
\end{equation}

The loss function of ACD, i.e. the binary cross entropy loss function of ACD, is given by:

\begin{equation}
\mathcal{L}_{ACD} = -\sum_{i=1}^{m} y_i^c \log p_i^c + (1-y_i^c)\log (1-p_i^c).
\end{equation}

\noindent
\textbf{Aspect-category sentiment classification}. The sentiment representations are obtained by $r_s = [\hat{\mathbf{g}}, \mathbf{u}_s]$. Thus, the sentiment probability $p_j^s \in \mathbb{R}^p $ corresponding to its category is given by:

\begin{equation}
p^s_j = p(y_j^s|y_j^c, r_j) = \mathrm{softmax}(\mathbf{W}^s_jr_s+\mathbf{b}^s).
\end{equation}

Following \citet{cai2020aspect}, a hierarchical prediction strategy is utilized to obtain the final sentiments prediction on the $i$-th category. We utilize cross-entropy loss as the objective loss function of ACSC which is defined as follows:

\begin{equation}
\mathcal{L}_{ACSC} = -\sum_{i=1}^m \sum_{j=1}^{p} \mathbb{I}(y_{i,j}^{s}) \log p^s_{i,j},
\end{equation}

\noindent
where $\mathbb{I}(\cdot)$ refers to an indicator function to adjust the output of the hierarchical prediction. The final loss is given by:

\begin{equation}
\begin{aligned}
  \mathcal{L}_{total} =  \delta_1 \mathcal{L}_{cl} + \delta_2 \mathcal{L}_{ACD} + \delta_3 \mathcal{L}_{ACSC}, \\
\end{aligned}
\end{equation}

\noindent
where $\delta_1, \delta_2, \delta_3 \in [0,1]$ are hyperparameters used to balance the three tasks, CL, ACD, and ACSC.

\begin{table}[t]
\begin{center}
\resizebox{1\linewidth}{!}{ 
\begin{tabular}{|l|r|r|r|r|}
\hline
      & REST 15 & REST 16 & LAP 15 & LAP 16 \\
      \hline
      \hline
\# Train sentences &   1102     &   1680     &   1397     & 2037       \\
\# Test sentences  &    572    &     580   &    644    &   572 \\ 
\# Categories  &    6    &     6   &    22    &   22 \\ 
\# Positive &   1458     &   1966     &  1641      &  2113  \\
\# Neutral  &    91    &    137    &   185    & 234   \\
\# Negative  &   663     &    848    &    1091     &  1353  \\
\hline
\end{tabular}
}
\caption{\label{datastatistics} Statistics of four benchmark datasets, REST and LAP in the SemEval-2015 and 2016.}
\end{center}
\end{table}

\section{Experiments}

\subsection{Datasets and Evaluation Metrics}

We evaluate our model on four benchmark datasets: REST 15 and LAP 15 from the SemEval-2015 task 12~\citelanguageresource{pontiki2015semeval}, and REST 16 and LAP 16 from the SemEval-2016 task 5 ~\citelanguageresource{pontiki2016semeval}. We choose these four datasets because the input of our ECAN is based not on a sentence unit but on a textual review to leverage document-level coherence. Each dataset consists of restaurant and laptop domains, and positive, neutral, and negative sentiment polarities. The statistics of datasets are displayed in Table~\ref{datastatistics}.  We use precision (P), recall (R), and macro-averaged F1 scores (F1) as metrics for evaluation.  

\begin{table*}[h]
\centering
\resizebox{1\linewidth}{!}{ 
\begin{tabular}{|c|ccc|ccc|ccc|ccc|}
\hline
& \multicolumn{3}{c|}{REST 15} & \multicolumn{3}{c|}{REST 16} & \multicolumn{3}{c|}{LAP 15} & \multicolumn{3}{c|}{LAP 16} \\
& P      & R      & F1      & P      & R      & F1      & P      & R      & F1      & P      & R      & F1 \\
\hline

CAER-BERT & 85.19	&76.42	&79.02	&84.81	&80.08	&82.14	&83.77	&62.86	&72.20	&81.09	&64.91	&72.67 \\ \hline
SCAN    &  86.23	&81.15	&83.62	&85.01	&85.01	&84.45	&80.34	&71.63	&74.82 	&81.39	&62.01	&71.12\\
AC-MIMLLN & \underline{88.64}	&80.76	&84.17	&86.05	&83.16	&82.59	&81.21	&69.36	&73.21
&82.09	&67.08	&74.22
\\
LC-BERT  & 87.62	&79.28	 &82.92	 &89.11	 &\underline{85.71} &85.40	&85.86	&65.25	&74.15	&82.57	&63.17	&70.85\\
EDU-Capsule  &84.63	 &78.42	 &81.40	 &89.91 	&84.21	&84.62	&92.47	&60.34	&73.03	&81.65	&63.32	&72.07 \\ \hline
Hier-BERT & 86.59	&\underline{80.86}	&83.59	&90.53	&85.23	&85.66	&88.27	&60.88	&72.06	&82.68	&66.38	&73.19\\
Hier-GCN-BERT &88.18	&79.28	&84.12	&\underline{91.16}	&85.29	&86.54	&\underline{92.53}	&75.10	&\underline{86.81}	&\underline{82.88} &74.42	&77.84\\
Hier-GCN-Xlnet & 88.19	&77.27	&\underline{84.39}	&90.44	&85.41	&\underline{87.26}	&91.78	&\underline{75.29}	&86.09	&82.16	&\underline{75.99}	&\underline{78.52} \\
\hline
Ours(ECAN) & \textbf{91.62} & \textbf{82.39} &\textbf{85.67} & \textbf{91.69} &\textbf{86.02} &\textbf{88.75}	&\textbf{93.64}	&\textbf{78.43}	&\textbf{88.80}	&\textbf{83.45}	&\textbf{76.23}	&\textbf{79.21} \\
\hline
\end{tabular}
}
\caption{ \label{result_acd} The main results of the ACD task.
%
Bold font and underline indicate the best, and the second best result, respectively.}
\end{table*}

\subsection{Baselines}

To examine the efficacy of ECAN in ACD and ACSC tasks, we compared it with the following nine SOTA baselines which are classified into four groups:

\vspace{0.5em}
\noindent
- \textit{\textbf{Single task learning-based}}: 
\vspace{0.5em}

\begin{itemize}[nosep,labelindent=0em,leftmargin=1em,font=\normalfont]
\item \textbf{CAER-BERT} \citep{liang2019context} adopts a sparse coefficient vector to select
highly correlated words from the sentences to adjust the representations of the aspect categories and sentiments for ACSA.
\end{itemize}

\vspace{0.5em}
\noindent
- \textit{\textbf{Multi-task learning-based}}:
\vspace{0.2em}

\begin{itemize}[nosep,labelindent=0em,leftmargin=1em,font=\normalfont]
\item \textbf{SCAN} \citep{li2020sentence} employs graph attention networks to aggregate high-order representations of the nodes in sentence constituency parse trees.
\vspace{0.5em}

\item \textbf{AC-MIMLLN} \citep{li2020multi} utilizes a multi-instance multi-label
learning network to obtain the sentiments of the sentence toward the aspect categories by aggregating the key instance sentiments.
\vspace{0.5em}

\item \textbf{LC-BERT} \citep{ wu2021locate} proposes a two-stage strategy that first locates the aspect term and then takes it as the bridge to find the related sentiment words.
\vspace{0.5em}

\item \textbf{EDU-Capsule} \citep{lin2023capsule} learns elementary discourse unit representations by capsule network within its sentential context.
\vspace{0.5em}
\end{itemize}

\noindent
-  \textbf{\textit{Hierarchical learning-based}}: 
\vspace{0.2em}

\begin{itemize}[nosep,labelindent=0em,leftmargin=1em,font=\normalfont]

\item \textbf{Hier-BERT} \citep{cai2020aspect} introduces the hierarchy method for ACD and ACSC tasks with BERT as the sentence encoder.
\vspace{0.5em}

\item \textbf{Hier-GCN-BERT} \citep{cai2020aspect} utilizes the GCN sub-layer to model the inner-relations between category and inter-relations between category and sentiment based on Hier-BERT.
\vspace{0.5em}

\item \textbf{Hier-GCN-XlNet} replaces the BERT encoder with a XLNet model in Hier-GCN-BERT model for comparison.
\vspace{0.5em}
\end{itemize}

\vspace{0.5em}
\noindent
-  \textbf{\textit{Generative approach-based}}:
\vspace{0.2em}

\begin{itemize}[nosep,labelindent=0em,leftmargin=1em,font=\normalfont]
\item \textbf{MvP} \citep{gouetal2023mvp}  introduces element order prompts to guide the language model to generate multiple sentiment tuples.
\vspace{0.5em}
\end{itemize}

\begin{table*}
\centering
\resizebox{1\linewidth}{!}{ 
\begin{tabular}{|c|ccc|ccc|ccc|ccc|}
\hline
Methods& \multicolumn{3}{c|}{REST 15} & \multicolumn{3}{c|}{REST 16} & \multicolumn{3}{c|}{LAP 15} & \multicolumn{3}{c|}{LAP 16} \\
& P  & R  & F1  & P & R & F1 & P& R  & F1 & P  & R & F1 \\
\hline
CAER-BERT & 71.27	& 63.75	& 66.25	& 77.69	& 68.39	& 73.60	& 58.34 & 51.86	&60.78	&65.36	&54.73	&60.79 \\ \hline
SCAN & 71.25	&67.05	&69.09	&73.87	&76.25	&75.05	&71.01	&45.49	&55.45	&61.06	&45.42	&52.08\\
AC-MIMLLN & 71.45 &65.18 &68.17	&75.66	&74.89	&75.27	&59.06	&63.78	&55.29	&68.08	&55.89	&61.39 \\
LC-BERT &  72.08	&65.76	&68.77	&76.34	&74.16	&75.75	&66.84	&50.80	&57.72	&\underline{71.05}	&53.11	&55.58 \\
EDU-capsule &72.67	&67.34	&69.90	&75.93	&\underline{77.00}	&74.81	&\underline{73.98}	&52.27	&65.43	&67.02	&49.20	&58.01 \\  \hline  
Hier-BERT &70.42	&65.75	&68.01	&76.99	&73.88	&75.41	&70.19	&48.41	&57.30	&66.01	&53.71	&59.23 \\
Hier-GCN-BERT  &75.12	&66.47	&\underline{71.94}	&77.62	&75.61	&\underline{77.68}	&75.00 &	\underline{66.84}	&\underline{70.69}	&68.32	&62.54	& 65.09\\
Hier-GCN-Xlnet & \underline{76.99}	&68.44	&70.53	&\underline{77.81}	&75.84	&76.35	&73.09	&64.17	&69.38	&70.3	&\underline{63.69}	&\underline{67.12} \\ \hline   
MvP* & 67.80 & \underline{68.63} & 68.21 & 73.76 & 75.49 & 74.62 & - & - & -  & - & - & - \\
\hline
Ours(ECAN) &\textbf{84.38}	&\textbf{74.43} &\textbf{79.09} & \textbf{84.92}	&\textbf{79.65}	&\textbf{82.20}	&\textbf{83.56}	& \textbf{75.34}	&\textbf{79.24}	&\textbf{76.11}	&\textbf{65.45}	& \textbf{70.45} \\
\hline
\end{tabular}
}
\caption{\label{result_acsa} The main results of the ACSC task. ``*'' refers to the results based on sentiment tuple prediction according to the original paper \cite{gouetal2023mvp}. }
\end{table*}

\begin{table*}[t]
\centering
\resizebox{1\linewidth}{!}{ 
\renewcommand\arraystretch{1.1}  
\begin{tabular}{|c|ccc|ccc|ccc|ccc|}
\hline
& \multicolumn{3}{c|}{REST 15} & \multicolumn{3}{c|}{REST 16} & \multicolumn{3}{c|}{LAP 15} & \multicolumn{3}{c|}{LAP 16} \\
& P      & R      & F1      & P      & R      & F1      & P      & R      & F1      & P      & R      & F1 \\
\hline
\hline
\multicolumn{13}{|c|}{ACD task}\\
\hline
\hline
w/o Senti-dis	&90.89	&80.13	&84.43	&90.78	&84.55	&86.80	&93.08	&76.19	&87.40	&82.59	&72.41	&76.85 \\
w/o Cate-dis	&89.87	&\textit{79.68}	&\textit{83.01} &90.55	&\textit{84.27}	&86.33	&92.06	&\textit{75.57}	&\textit{86.76} &\textit{81.79}	&\textit{70.12}	&\textit{75.32} \\
w/o W-syn	&89.67	&80.23	&84.11	&91.20	&84.62	&87.85	&92.90	&77.19	&87.80	&82.98	&73.17	&77.76 \\
w/o cl & \textit{87.95}	&79.83	&84.19	&\textit{88.77}	&85.57	&\textit{86.01}
&\textit{91.96}	&76.32	&87.31	&83.08	&71.86	&77.31\\
\hline
Ours(ECAN)      & \textbf{91.62} & \textbf{82.39} &\textbf{85.67} & \textbf{91.69} &\textbf{86.02} &\textbf{88.75}	&\textbf{93.64}	&\textbf{78.43}	&\textbf{88.80}	&\textbf{83.45}	&\textbf{76.23}	&\textbf{79.21} \\
\hline
\hline
\multicolumn{13}{|c|}{ACSC task}\\
\hline
\hline
 w/o Senti-dis &	79.71	&\textit{72.76}	&\textit{74.07}	&83.17	&77.34	&79.12	&82.18	&\textsl{72.23}	&\textit{76.05}	&73.83	&63.12	&68.89  \\
w/o Cate-dis &82.59	&74.15	&75.14	&84.49	&79.37	&81.84	&83.35	&72.60	&76.42	&74.91	&\textit{62.68}	&69.09 \\
w/o W-syn &	\textit{79.28}	&73.30	&75.63	&82.54	&77.78	&80.58	&\textit{81.60}	&72.35	&77.76	&74.72	&63.72	&68.74 \\
w/o cl &80.92	&72.86	&75.25&	\textit{82.09} &\textit{76.76} &\textit{79.01}
 &82.27	&72.86 &77.28	&\textit{72.87}	& 63.39	&\textit{67.05} \\
\hline
Ours(ECAN) &\textbf{84.38}	&\textbf{74.43} &\textbf{79.09} & \textbf{84.92}	&\textbf{79.65}	&\textbf{82.20}	&\textbf{83.56}	& \textbf{75.34}	&\textbf{79.24}	&\textbf{76.11}	&\textbf{65.45}	& \textbf{70.45} \\
\hline
\end{tabular}
}
\caption{ \label{result_ablation_study} Ablation study. ``w/o Senti-dis'' refers to the result without the sentiment disentanglement part, ``w/o Cate-dis'' indicates the result without the category disentanglement part, ``w/o W-syn'' shows the result without word-level syntactic learning, and ``w/o cl'' stands for the result without the contrastive learning loss. The italic font value indicates the worst results.}
\end{table*}

\subsection{Implementation Details}

Following \citet{cai2020aspect}, we randomly chose 10\% of the training data and used it as the development data. The optimal hyper-parameters were as follows: The initial learning rate for coherence-aware representation learning was 8e-6, and the others were 2e-5. The weight decay was set to 1e-3, and the dropout rate was 0.1. The number of negative samples $B$ was 5, and the margin $\tau$ was 0.1. The balance coefficients $\delta_1$, $\delta_2$, and $\delta_3$ were set to 0.1, 0.5, and 0.5, respectively. The number of graph convolutional layers was 3. The number of disentangled blocks was set to 4 for both category disentanglement $d_c$ and sentiment disentanglement $d_s$. All hyperparameters were tuned using Optuna\footnote{https://github.com/pfnet/optuna. 
The search ranges are reported in the appendix~\ref{sec:hyperparameters}}.
We used AdamW \cite{loshchilov2017decoupled} as the optimizer.

\subsection{Main Results}

The experimental results on the four datasets show that the ECAN exceeds all baselines in both ACD and ACSC tasks, indicating the effectiveness of disentanglement of category and sentiment for coherence-aware representations.
Specifically, it surpasses baselines by a large margin in the ACSC task, while being slightly better than the second-best methods in the ACD task. 

\subsubsection{Performance of Aspect Category Detection}
\begin{figure*}[t]
\begin{center}
\includegraphics[width=\linewidth]{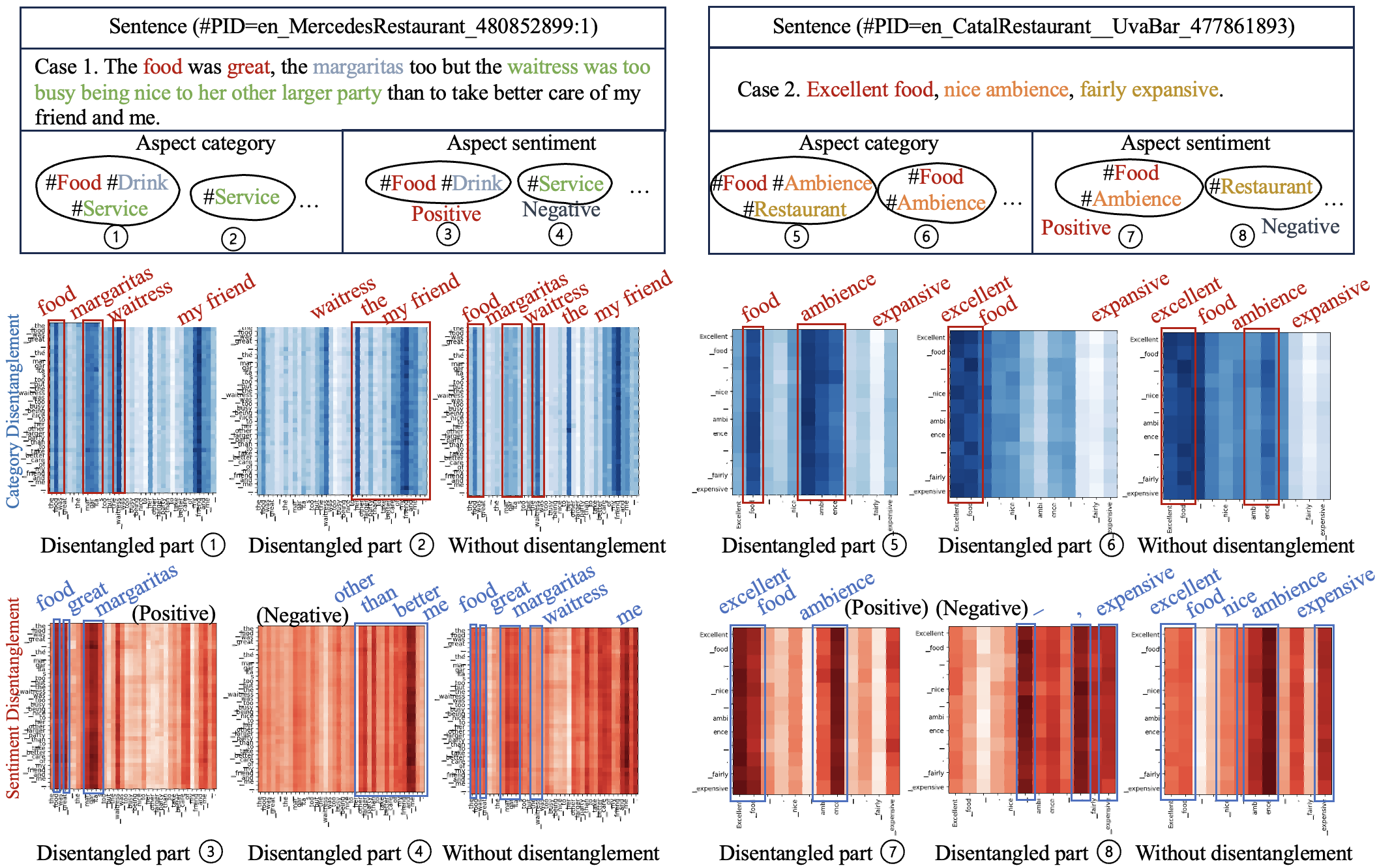} 
\caption{The visualization result of category and sentiment disentanglement. The top box on the left/right side indicates a sentence in a review with the source and disentangled categories and sentiments by the ECAN. The blue color and the red color denote the visualized correlations between categories and between sentiments in disentanglements. The warmer the tone colors, the higher the correlations.}

\label{vis_fig}
\end{center}
\end{figure*}

Table~\ref{result_acd} shows the results of aspect category detection. Overall, the ECAN attained an improvement over the second-best methods by 0.6$\sim$3.4\% in precision,  0.3$\sim$4.2\% in recall, and 0.9$\sim$2.3\% in the F1-score. Table~\ref{result_acd} also prompts the following observations and insights:
\vspace{0.5em}
\begin{itemize}[nosep,labelindent=0em,leftmargin=1em,font=\normalfont]
\item Hierarchical learning, such as Hier-GCN-BERT and Hier-GCN-XLNet, are competitive among baselines, while the single task learning-based method shows the worst results in all datasets. One reason is that hierarchical learning approaches build hierarchical category-sentiment graphs to aggregate the inter-relationships between categories and sentiments by co-occurring category pairs, which balance the effect of high-frequency and low-frequency categories. 
\vspace{0.5em}
\item In addition to modeling the hierarchical relations between categories and sentiments like hierarchical learning, the ECAN also learns the document-level contexts to assist in the category detection task via coherence-aware contrastive learning, which helps to achieve superior performance.
\vspace{0.5em}
\item Interestingly, sentiment disentanglement is also beneficial for identifying the categories in the sentences. As we can see from Figure~\ref{vis_fig}, the aspect term, e.g., ``food,'' is also highlighted when identifying the positive sentiment word, e.g., ``excellent,'' in the sentiment disentanglement part \ding{178}. It supports that the sentiment disentanglement process helps to connect the relations between sentiments and categories due to the shared representations optimized by multi-task learning.
\end{itemize}

\subsubsection{Performance of Aspect Category Sentiment Classification}

Table~\ref{result_acsa} shows the results of aspect category sentiment classification. Overall, the ECAN attained an improvement over the second-best methods by 7.1$\sim$11.4\% in precision,  2.8$\sim$12.7\% in recall, and 5.0$\sim$12.1\% in the F1-score. In particular, it achieved remarkable results for REST 15 and LAP 15 compared with all baselines. The improvement of 7.8\% on average through all metrics in the ACSC task is significantly better than that of 1.6\% in the ACD task. In addition to similar observations in the ACD task, we have the following findings:
\begin{itemize}[nosep,labelindent=0em,leftmargin=1em,font=\normalfont]
\vspace{0.5em}
\item The ECAN significantly outperforms all baseline models, highlighting its capability to achieve superior performance by disentangling sentiment representations related to aspect categories. Furthermore, coherence-aware representation learning and word-level syntactic learning play crucial roles in enhancing sentiment semantics.
\vspace{0.5em}
\item The Hier-BERT and LC-BERT work well in the ACD task, while they fail to effectively identify the aspect-category sentiment, as they do not utilize GCN to learn many-to-many relatedness between aspect categories and sentiments for node embedding enhancement.

\end{itemize}

\subsection{Ablation Study}

We conducted ablation experiments to verify the effectiveness of each component in the ECAN. As shown in Table~\ref{result_ablation_study}, the results in the ACD and ACSC tasks prompt the following observations:
\vspace{0.5em}
\begin{itemize}[nosep,labelindent=0em,leftmargin=1em,font=\normalfont]
\item As expected, category disentanglement (i.e., w/o Cate-dis) contributes the most to the ACD task, and sentiment disentanglement (i.e., w/o Senti-dis) works better in the ACSC task. It also supports that without hierarchical disentanglement, the performance is fairly limited by the entangled sentiments within each sentence.
\vspace{0.5em}

\item Without coherence-aware representation by contrastive learning (i.e., w/o cl), the performance of the ECAN significantly drops in both the ACD and ACSC tasks, demonstrating that document-level coherence is a strong clue in assisting in learning coherence-aware representation.
\vspace{0.5em}

\item The underperformance of the ECAN without word-level syntactic learning (w/o W-syn) in the ACSC task, indicates that syntactic dependencies help to derive enhanced sentiment semantic representations. We also found that syntactic analysis contributes to the ACSC of sentences containing conflicting sentiment expressions, such as the sentence ``Even the chickpeas, which I normally find too dry, were good''.

\end{itemize}

\subsection{Visualization of Hierarchical Disentanglement}

To understand how the hierarchical disentanglement model assists in ACSA, we present two cases and provide performance comparisons through visualization. The comparisons are (i) the disentangled category and sentiment parts in reviews, and (ii) the counterparts without disentanglement.

\noindent
\textbf{Case 1.} There are three aspect categories, ``food,'' ``drink,'' and ``service,'' and their sentiments are positive, positive, and negative, respectively. From the left part of Figure~\ref{vis_fig}, we make the following observations:

\begin{itemize}[nosep,labelindent=0em,leftmargin=1em,font=\normalfont]
\vspace{0.5em}

\item For category disentanglement, the disentangled part \ding{172} captures the aspect categories of ``food,'' ``drink,'' and ``service,'' and part \ding{173} disentangles ``service'' from others;
\vspace{0.5em}

\item For the sentiment disentanglement, part \ding{174} detected positive sentiment of aspect categories, ``food'' and ``drink'' and part \ding{175} highlighted the words expressing negative sentiments, as the review mentions the waitress put more efforts on her other larger party;
\vspace{0.5em}

\item  Without disentanglement, the model could focus on only one part of the categories, i.e., it highlighted the negative sentiment on the waitress while reducing the impact of the sentiments on food and drink.
\vspace{0.5em}

\end{itemize}

\noindent
\textbf{Case 2.} Three aspect categories are mentioned in the review, i.e., ``food,'' ``ambiance,'' and ``restaurant,'' with positive, positive, and negative sentiment polarities, respectively. By observing the right part of  Figure~\ref{vis_fig}, we have the following findings:

\begin{itemize}[nosep,labelindent=0em,leftmargin=1em,font=\normalfont]
\vspace{0.5em}

\item For category disentanglement, the disentangled part \ding{176} recognizes the categories of ``food'' and ``ambiance,'' while part \ding{177} concentrates on ``food'';
\vspace{0.5em}

\item For the sentiment disentanglement, part \ding{178} correctly identifies the positive sentiments as it focuses on ``excellent food'' and ``ambiance.'' In contrast, part \ding{179} avoided the attention of positive words, such as ``excellent" and ``nice'' and focused on ``expensive'' and other neutral tokens, such as ``-'' and ``' ''. 
\vspace{0.5em}

\end{itemize}

\subsection{Error Analysis}
We conducted error analyses on the REST 16 and LAP 16 datasets. 
There are two major types of errors for the ACD task:

\begin{itemize}[nosep,labelindent=0em,leftmargin=1em,font=\normalfont]
\vspace{0.5em}
\item Generalized categories, such as ``restaurant\# general'' in the restaurant domain and ``laptop\# general'' in the laptop domain, pose difficulties in our ECAN. For example, the category entity of \textit{``great lunch spot.''} should be ``restaurant,'' while the ECAN predicts it to be ``food'' incorrectly.
\vspace{0.5em}
\item The special terms that appear with low frequency, such as names of places, persons, and products, are often ignored by our ECAN. This inspires us to inject more knowledge, e.g., by leveraging the large language model \cite{zhang2023sentiment}. This remains a rich space for further exploration.
\vspace{0.5em}
\end{itemize}

\noindent
We also found two typical cases of errors for the ACSC task:

\begin{itemize}[nosep,labelindent=0em,leftmargin=1em,font=\normalfont]
\vspace{0.5em}
\item  In the case of a sentence with more than two transitions, for example, \textit{``I liked the atmosphere very much but the food was not worth the price.''}, the ECAN incorrectly predicts positive polarity on the price of the food. It inspired us to explore special feature extractors for this to further improve the performance.
\vspace{0.5em}
\item  Neutral sentiment is still challenging for the ECAN, although we observed that it identified positive and negative sentiments well. One of the reasons is that the shortage of neutral training examples, which is shown in Table~\ref{datastatistics}, complicates its identification. Analyzing neutral sentiments is also an interesting direction for future work.

\end{itemize}




\section{Conclusion} 

We proposed an enhanced coherence-aware network (ECAN). It leverages coherence modeling to learn explicit opinions from contexts and a hierarchical disentanglement of category and sentiment to mine more fine-grained features for both ACD and ACSC tasks.
Extensive experimental and visualization results show that our ECAN effectively decouples categories and sentiments entangled in the coherence representations and achieves SOTA performance. 

Future work includes (1) seeking more interpretable solutions to disentangle the categories and sentiments; and (2) exploring stronger clues to guide the disentangled channel to learn independent features between channels, such as maximizing the variances of various disentangled representations.

\section*{Limitations}
There is space for further improving the ability of ECAN to disentangle categories. 
The ECAN model adopts time-consuming modules, i.e., XLNet ($O(n^2)$) and GCN ($O(n^2)$) where $n$ refers to the number of words, therefore its computational cost heavily relies on the length of textual reviews.

\section*{Ethics Statement}
This paper does not involve the presentation of a new dataset, an NLP application, and the utilization of demographic or identity characteristics information.

\section*{Acknowledgements}
We would like to thank anonymous reviewers for their thorough comments and suggestions. This work is supported by JKA and Kajima Foundation’s Support Program and the China Scholarship Council (No.202208330091).

\normalem
\section{Bibliographical References}\label{sec:reference}
\bibliographystyle{lrec-coling2024-natbib}
\bibliography{ECAN_bib_example}

\section{Language Resource References}
\label{lr:ref}
\bibliographystylelanguageresource{lrec-coling2024-natbib}
\bibliographylanguageresource{languageresource}

\appendix
\section{Appendix}

\subsection{Implementation and hyperparameter setting}
\label{sec:hyperparameters}

We implemented ECAN and had experimented with Pytorch on a single GPU: NVIDIA GeForce RTX 3090 (24GB memory). The search ranges of the hyperparameters used in our experiments are shown in Table \ref{tab:hyperparameters}.

\begin{table}[h]
\scalebox{0.8}{
\begin{tabular}{c|c}
\hline
Parameter & Range \\ \hline
LR of coherence modeling & 1e-6 $\sim$ 1e-5 \\ \hline
LR of others & 1e-5 $\sim$ 1e-4  \\ \hline
Weight decay & \{1e-4, 1e-3, 1e-2\}\\\hline
Dropout rate & \{0.1, 0.2, 0.3\} \\ \hline
\#Negative Samples $B$  & \{5, 6, 7, 8, 9, 10\} \\ \hline
Margin $\tau$  & \{0.05, 0.1, 0.15, 0.2\} \\ \hline
$\delta_1$ & \{0.05, 0.1, 0.15, 0.2\} \\ \hline
$\delta_2$ & \{0.3, 0.4, 0.5, 0.6, 0.7, 0.8\} \\ \hline
$\delta_3$ & \{0.3, 0.4, 0.5, 0.6, 0.7, 0.8\} \\ \hline
\#Block of GCN & \{1, 2, 3\} \\ \hline
\# disentangled blocks & \{1, 2, 3, 4, 5, 6, 7\} \\ \hline
\end{tabular}
}
\caption{Search range of each hyperparameter: LR refers to the learning rate. LR of coherence modeling indicates the learning rate of coherence-aware representation learning. LR of others shows hierarchical disentanglement and word-level syntactic learning.}  
\label{tab:hyperparameters}
\end{table}

\begin{figure*}[h]
\begin{center}
\includegraphics[width=\linewidth]{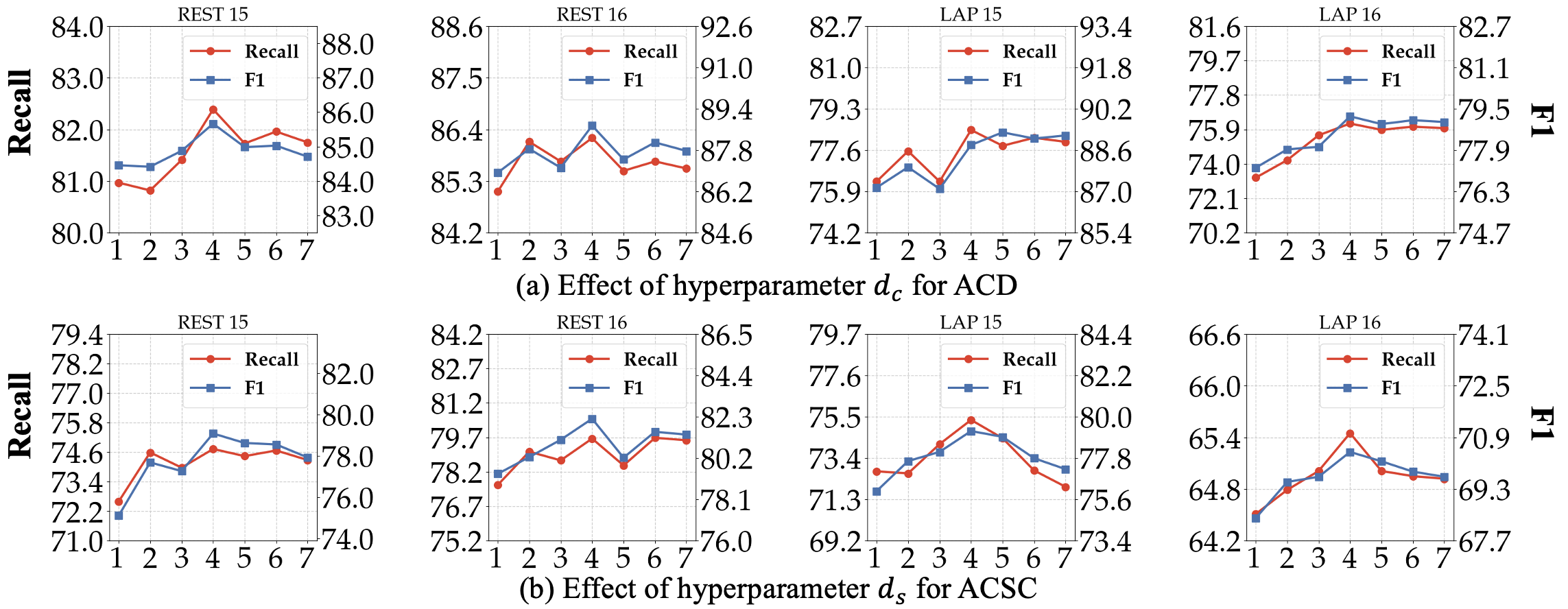} 
\caption{The effect of hyperparameters $d_c$ and $d_s$.}
\label{para_fig}
\end{center}
\end{figure*}

\subsection{Parameter Analysis }
We examined how key hyperparameters, the number of category disentanglement blocks $d_c$, and the number of sentiment disentanglement blocks $d_s$ affected the performance of the ECAN. The results are given in Figure~\ref{para_fig}.

\noindent
\textbf{Effect of $d_c$.} The highest value of $d_c$ is four for all datasets. We can also observe that on the REST 15 and REST 16 datasets, the performance is better when $d_c$ increases to four, while the performance decreases when $d_c$ becomes higher than four. On LAP 15 and LAP 16, the performance does not improve. However, its computational cost increases with $d_c$ being higher than four.

\noindent
\textbf{Effect of $d_s$.} The best value of $d_s$ is also four for all datasets. It is reasonable that the performance is getting worse with $d_c$ being higher than four because there are only three types of sentiment polarities and a higher number of disentanglement parts could mislead the model.

\end{document}